\documentclass[conference]{IEEEtran}
\renewcommand{\baselinestretch}{0.9602}
\usepackage{multicol}
\usepackage{multirow}
\usepackage{caption}
\usepackage{graphicx}
\usepackage{subfig}

\begin{document}

\title{Global Model Interpretation via\\ Recursive Partitioning}

\author{\IEEEauthorblockN{Chengliang Yang}
	\IEEEauthorblockA{\textit{Department of Computer \&}\\
		\textit{Information Science \& Engineering}\\
		\textit{University of Florida}\\
		Gainesville, Florida 32611\\
		Email: ximen14@ufl.edu}
	\and
	\IEEEauthorblockN{Anand Rangarajan}
	\IEEEauthorblockA{\textit{Department of Computer \&}\\
		\textit{Information Science \& Engineering}\\
		\textit{University of Florida}\\
		Gainesville, Florida 32611\\
		Email: anand@cise.ufl.edu}
	\and
	\IEEEauthorblockN{Sanjay Ranka}
	\IEEEauthorblockA{\textit{Department of Computer \&}\\
		\textit{Information Science \& Engineering}\\
		\textit{University of Florida}\\
		Gainesville, Florida 32611\\
		Email: ranka@cise.ufl.edu}}

\maketitle

\begin{abstract}
	In this work, we propose a simple but effective method to interpret black-box machine learning models globally. That is, we use a compact binary tree, the interpretation tree, to explicitly represent the most important decision rules that are implicitly contained in the black-box machine learning models. This tree is learned from the contribution matrix which consists of the contributions of input variables to predicted scores for each single prediction. To generate the interpretation tree, a unified process recursively partitions the input variable space by maximizing the difference in the average contribution of the split variable between the divided spaces. We demonstrate the effectiveness of our method in diagnosing machine learning models on multiple tasks. Also, it is useful for new knowledge discovery as such insights are not easily identifiable when only looking at single predictions. In general, our work makes it easier and more efficient for human beings to understand machine learning models.
\end{abstract}


\section{Introduction}
Though machine learning advances greatly in many areas in recent years such as computer vision and natural language processing, limited interpretability hinders it from impacting areas that require clearer evidence for decision making such as health care and economy. In these domains, most widely used machine learning models are linear regression or decision trees that people can easily understand. To deploy cutting-edge machine learning in such domains, some transparent mechanisms are needed to explain the sophisticated models to users. Limited interpretability also harms improving machine learning models. The black-box behavior makes it difficult to diagnose the models. Without good understanding of how the model works, lots of effort are wasted in model parameters tuning.

Thus, machine learning researchers have been trying to better interpret machine learning models. Recent progresses include designing specific neural network structure that imposes linear constraints on weights of input variables in the decision function \cite{choi2016retain}, using model structure based heuristics to decompose prediction scores \cite{yang2016predicting}, approximating any model linearly in a local area \cite{ribeiro2016should}, and track predictions using influence functions back to training data \cite{koh2017understanding}. However, all these work can only provide local interpretation. That is, the interpretation is generated for a particular sample of data. This is not desired when we want to know the general picture of the model. Modern machine models are usually trained and tested on millions of data samples. It is not practical for a researcher to review the interpretation of each of them for model diagnostics. What's more, when machine learning models are used to inform population level decisions, such as an economic policy change, a global effect estimate would be more helpful than thousands of local explanations. Thus there is a gap between local and global machine learning model interpretation, which this paper is going to address.

By "interpreting a machine learning model globally", we mean representing a trained machine learning model in an aggregated and human understandable way. This is done by extracting the most important rules that the model learned from training data and would apply to testing data. These rules affect a substantial portion of data from the model perspective and thus are useful to inform decision impacting globally for all data samples. The simplest example of such rules are the coefficients we could learn in a linear regression model. These coefficients represent the magnitude of changes in the output due to one unit of change in the input variables. From the assumption of the linear model, the coefficients are identical for any data sample. Thus they are used widely for effect estimation in observational studies and randomized experiments. Another globally interpretable machine learning model is decision tree, which presents the decision rules in a straightforward tree structure. However, both linear regression and decision tree lack high predictive power. which means people have to tradeoff between predictability and interpretability when both properties are desired.

\begin{figure}[!t]
	\centering{} 
	\includegraphics[width=80mm]{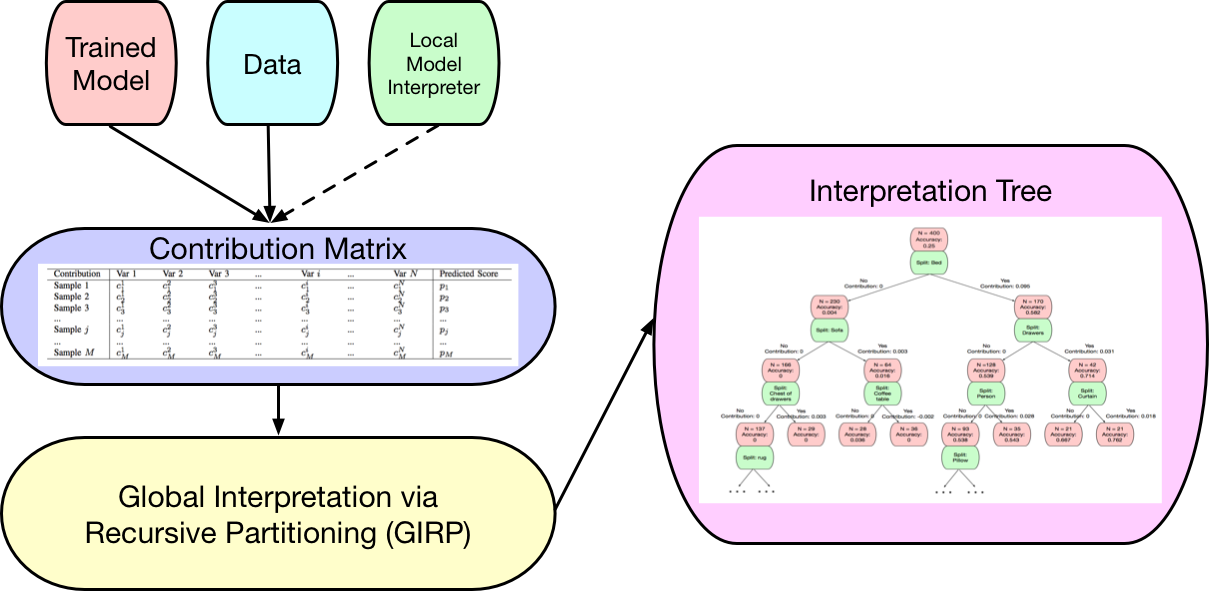} \caption{Work flow of global model interpretation}
	\label{flow} 
\end{figure}

In this work, we propose a new method, Global Interpretation via Recursive Partitioning (GIRP), to build a global interpretation tree for a wide range of machine learning models based on their local explanations. That is, we recursively partition the input variable space by maximizing the difference in the contribution of input variables averaged from local explanations between the divided spaces. By doing so, we end up with a binary tree that we call the interpretation tree describing a set of decision rules that is an approximation of the original machine learning model. Figure \ref{flow} describes the work flow of building a global model interpretation. With a trained machine learning model and the data you want to use to explain it, we first generate a contribution matrix from local explanations either using model specific heuristics or some local model interpreter \cite{ribeiro2016should}. Then we send contribution matrix to our Global Interpretation via Recursive Partitioning (GIRP) algorithm. The algorithm returns with an interpretation tree that generally describes the machine learning model and is fully comprehensible by human beings. The key contributions of our paper are as follows:
\begin{itemize}
	\item
	We propose an efficient and effective method to address the gap in the literature of globally interpreting many machine learning models. The interpretation is in the form of an easily understandable binary tree that could be used to diagnose the interpreted model or inform population level decisions.
	\item
	The CART \cite{cart84} like algorithm that we use to build the interpretation tree could model interactions between input variables. Thus, we are able to find the heterogeneity in variable importances among different subgroups of data.
	\item
	In our experiments, we showcase that our method can discover whether a particular machine learning model is behaving in a reasonable way or overfit to some unreasonable pattern.
\end{itemize}
The rest of the paper is organized as follows. Section 2 describes some works that are closely connected to our work. Section 3 presents the Global Interpretation via Recursive Partitioning (GIRP) algorithm. Section 4 applies GIRP to computer vision, natural language processing, and health care predictive models from structured tabular data. Section 5 concludes the paper.

\section{Related Work}
There are four parts of existing work that are closely related to our method, that is, local model interpretation, global model interpretation, recursive partitioning for effects estimation, and feature selections.

There are several ways to achieve local model interpretation. First, people structure the model in a way that the output is linear in terms of input variables so that the weights could be used as a measure of importance. For example, \cite{choi2016retain} uses the neural attention mechanism \cite{bahdanau2014neural} to generate interpretable attention weights in recurrent neural networks. However, due to the stochastic training process used, these attentions are shown to be unstable \cite{yang2017machine}. The second way uses model specific heuristics to decompose the predicted scores by input variables. \cite{yang2016predicting} described such methods for regularized regression and gradient boosted machine. Third, locally approximating sophisticated models using simple interpretable model could explain individual predictions. Gradient vector and sparse linear methods are tried as the local explainer \cite{kononenko2010efficient,baehrens2010explain,ribeiro2016should,yang2018visual}. Finally, influence functions from robust statistics can be used to track a particular prediction back to training data that are responsible for it \cite{koh2017understanding}. In conclusion, all local model interpretation methods work at the single data sample level, generating the contributions of input variables to the final predicted score for a specific data sample.

People also try to directly build a globally interpretable model, including additive models for predicting pneumonia risk \cite{caruana2015intelligible} and rule sets generated from sparse Bayesian generative model \cite{letham2015interpretable}. However, these models are usually specifically structured thus limited in predictability to preserve interpretability.  \cite{craven1996extracting} uses queries to build a tree to approximate neural networks. \cite{atzmueller2011mining} generally discusses presenting machine learning models in different levels.

Recursive partitioning and its resulting tree structure is an intuitive way to present rule sets and model interactions between input variables. It has been used for a long time to analyze heterogeneity for subgroup analysis in survey data \cite{morgan1963problems}. Recently it is applied to study heterogeneous causal or treatment effects \cite{su2009subgroup,athey2016recursive}. It is a good fit for our global model interpretation task because we want to extract the rules that machine learning model finds and these rules are affected by the interactions between input variables.

Feature selection methods select a subset of important features from the input variables when the machine learning model is trained. The model interpretability could be benefited from this process because it reduces the dimension of input variables, making the model compact and easier to be presented \cite{tibshirani1996regression}. This is very useful when the input is very high dimensional \cite{mungloo2017meta}. The feature selection process could either be conducted before the model fitting \cite{hall1999correlation} or embedded into it \cite{tibshirani1996regression,xu2014gradient}. Though feature selection and global model interpretation tasks both aim to extract the most important variables or their combinations, they are different because global model interpretation is a post model fitting process. We represent the trained model in a compact and comprehensible way with good fidelity to the original model. The goal is not to make predictions using this representation but understand how it predicts. In contrast, feature selection discards unimportant variables and predictions will be solely based on selected ones.

\section{Global Interpretation via Recursive Partitioning}

We follow the CART \cite{cart84} work flow to build the interpretation tree, including growing a large initial tree, pruning, and using the validation set for best tree size selection. But before we describe the tree building process in detail, we introduce the contribution matrix first, which our method takes as input.

\subsection{Contribution Matrix}

As mentioned, local model interpretation methods \cite{kononenko2010efficient,baehrens2010explain,ribeiro2016should,choi2016retain,yang2016predicting} can generate the contribution of each single input variable to the final predicted score for a specific data sample. In detail, for a machine learning model that take $N$ input variables, given a new data sample, it generates a quantity $c^{i}$ for the $i$-th variable $v_{i}$ to measure the importance of this variable in the prediction made. We call this quantity the contribution of variable $v_{i}$. If there are $M$ data samples in total, we could generate a contribution matrix using local model interpretation methods as shown in Table \ref{comp}. $c_{j}^{i}$ is the contribution of variable $v_{i}$ to predicted score $p_{j}$ of sample $s_{j}$. Thus, each row of the contribution matrix represents how the model thinks of variable importances in the corresponding prediction.

\begin{table*}[t]
	\begin{center}
		\begin{tabular}{p{2.5cm}|p{1cm}p{1cm}p{1cm}p{1cm}p{1cm}p{1cm}p{1cm}|p{2cm}}
			\hline
			Contribution & Var 1 & Var 2 & Var 3 & ... & Var $i$ & ... & Var $N$ & Predicted Score\\
			\hline
			Sample 1 & $c_{1}^{1}$ & $c_{1}^{2}$ & $c_{1}^{3}$ & ... & $c_{1}^{i}$ &... & $c_{1}^{N}$ & $p_{1}$\\
			Sample 2 & $c_{2}^{1}$ & $c_{2}^{2}$ & $c_{2}^{3}$ & ... & $c_{2}^{i}$ &... & $c_{2}^{N}$ & $p_{2}$\\
			Sample 3 & $c_{3}^{1}$ & $c_{3}^{2}$ & $c_{3}^{3}$ & ... & $c_{3}^{i}$ &... & $c_{3}^{N}$ & $p_{3}$\\
			...      & ... & ... & ... & ... & ... & ... & ... & ... \\
			Sample $j$ & $c_{j}^{1}$ & $c_{j}^{2}$ & $c_{j}^{3}$ & ... & $c_{j}^{i}$ &... & $c_{j}^{N}$ & $p_{j}$\\
			...      & ... & ... & ... & ... & ... & ... & ... & ... \\
			Sample $M$ & $c_{M}^{1}$ & $c_{M}^{2}$ & $c_{M}^{3}$ & ... & $c_{M}^{i}$ &... & $c_{M}^{N}$ & $p_{M}$\\
			\hline            
		\end{tabular}
	\end{center}
	\caption{
		Contribution matrix generated from local model interpretations for every single data sample. $c_{j}^{i}$ is the contribution of variable $v^{i}$ to predicted score $p_{j}$ of sample $s_{j}$.}
	\label{comp}
\end{table*}

It is straightforward to obtain the contribution matrix when features are explicit and individual contributions could be generated along with predictions like in linear regression and \cite{choi2016retain,yang2016predicting}. However, in other cases we need some workaround to identify the variables that the contributions could be attributed to. When analyzing convolutional neural networks in \cite{ribeiro2016should}, segmentation of images is generated first to carry contributions. In our experiment diagnosing the scene classification deep learning method, a semantic segmentation algorithm is applied to the scene images to break them into semantic meaningful segments as well. These workarounds are problem specific and affect the formation of the contribution matrix.

\subsection{Growing a Large Initial Tree}
Now we can move forward to the first step to build the interpretation tree, growing a large initial tree. The same greedy process as CART is adopted. For any input variable $i$, we could apply a split based on values of variable $i$ to divide all the data samples into two subgroups. Note that the split is based on the input variable value but not contribution $c^{i}$. We use $v^{i}$ to denote the input values to discriminate it from contribution $c^{i}$. The type of split depends on type of variable $v^{i}$. If it is binary, the split criteria could be "$v^{i} = 1$?". If $v^{i}$ is ordinal, we could apply the criteria "if $v^{i} < d$" where $d$ is some constant value. If $v^{i}$ is categorical, let $D$ denote a subset of all possible values of variable $v^{i}$, we could apply "$v^{i} \in D$?" as the split criteria. For convenience, assume that all data samples meet the split criteria go to the right subset $S_{R}$ and the others go to the left subset $S_{L}$. For the two subsets of data samples $S_{R}$ and $S_{L}$. Consider the quantity below:

\begin{equation}
G(split_{i}) = \left(\frac{\sum_{S_{L}}c_{j}^{i}}{|S_{L}|} - \frac{\sum_{S_{R}}c_{j}^{i}}{|S_{R}|}\right)
\label{eq1}
\end{equation}

\noindent $split_{i}$ means the split is over variable $v^{i}$. The first term quantifies the average contribution of variable $v_{i}$ in the left subset $S_{L}$. So does the second term for the right subset $S_{R}$. The difference between these two terms measures how differently variable $v^{i}$ contributes to the predicted score in $S_{R}$ and $S_{L}$. The larger this difference is, the more discriminative the model think variable $v^{i}$ is. Thus, by finding the maximum $|G(split_{i})|$, we could get to know the most import variable from the model perspective. So $|G(split_{i})|$ is used as a measure of split strength in terms of variable importance.

We search all possible splits for all variables to find the best initial split. After dividing the data sample into $S_{R}$ and $S_{L}$, we follow CART's greedy approach to recursively partition $S_{R}$ and $S_{L}$ and their child nodes until we reach to some pre-set threshold for maximum tree depth or minimum number of samples in a node. As a result of this step, we would get a large initial tree that explicitly represents the most discriminative rules that the model implicitly contains. We denote this large initial tree $T_{0}$.

\subsection{Pruning}
Due to the greedy approach to grow the initial tree, the rules contained in initial tree $T_{0}$ are overly optimistic about the real world problem and may not generalize well. Thus we need a procedure to prune $T_{0}$ to improve generalizability. Consider all the internal nodes (non-leaf nodes) in $T_{0}$. All these nodes contain a split, say $t$. Each split corresponds to a split value $G(t)$ defined by Equation (\ref{eq1}). Suppose $T$ is any interpretation tree and $t$ is an internal node of $T$, we have

\begin{equation}
G(T) = \sum_{t \in T}|G(t)|
\end{equation}

\noindent as a measure of the total split strength of the tree $T$ that we generally want to maximize. To control the complexity of $T$, we add a penalizing term to $G(T)$ to punish for more nodes in the tree.

\begin{equation}
G_{\lambda}(T) = \sum_{t \in T}|G(t)| - \lambda|T|
\end{equation}

\noindent Here $|T|$ stands for the number of internal nodes in $T$. To maximize $G_{\lambda}(T)$, some of the internal nodes need to be removed from $T$ if $G(t)$ for these nodes are less than $\lambda$. For larger $\lambda$, more nodes would be removed so the resulting tree would be simpler and vice versa. But how can we decide which internal nodes to remove? We first define a new quantity for each internal nodes $t$ for this purpose. We use $T_{t}$ to denote the subtree of $T_{0}$ that has $t$ as root.

\begin{equation}
g(T_{t}) = \frac{|G(T_{t})|}{|T_{t}|}
\label{eq4}
\end{equation}

\noindent The above quantity intuitively defines the average split strength of internal nodes in subtree $T_{t}$. With $g(T_{t})$ defined, we iteratively remove the subtree with the smallest $g(T_{t})$ from the initial full tree $T_{0}$. Due to the greedy process to grow the initial tree, this process would result in a series of nested tree, $\{T_{K}, T_{K-1}, ..., T_{k}, T_{k-1}, ...,T_{0}\}$. $T_{K}$ is the null tree that only contains one node. \cite{cart84} has proved that these nested trees created by the iterative pruning process correspond to a series of $\lambda$ values, with $\lambda_{K} > \lambda_{K-1} > ... > \lambda_{k} > ... > \lambda_{0} = 0$.

\subsection{Select Best Sized Tree}
But how can we decide which $T_{k}$ is the best sized tree for the final interpretation tree, i.e., which value of $\lambda_{k}$ is the best? Here we use a held-out validation set for making this decision. We feed these new validation data into each of $\{T_{K}, T_{K-1}, ..., T_{k}, T_{k-1}, ...,T_{0}\}$ and calculate for each internal node $t$

\begin{equation}
G_{validation}(t) = sgn(G(t))\left(\frac{\sum_{S_{L}}c_{j}^{i}}{|S_{L}|} - \frac{\sum_{S_{R}}c_{j}^{i}}{|S_{R}|}\right)
\label{eq5}
\end{equation}

\noindent where $sgn()$ is the sign function. Then we select the tree $T_{k}$ as the best sized tree with the largest $G_{validation}(T_{k})$:

\begin{equation}
G_{validation}(T_{k}) = \sum_{t \in T_{k}}(G_{validation}(t))
\label{eq6}
\end{equation}

\subsection{Choice of Hyperparameters}
The only two hyperparameters we have in our approach are the maximum depth of the interpretation tree and the minimum number of data samples within a leaf or internal node. These are mostly chosen empirically depending on the problem setting. 

\begin{table}[!t]
	\begin{center}
		\begin{tabular}{lp{6.2cm}}
			\hline
			\textbf{Algorithm:} & Global Interpretation via Recursive Partitioning (GIRP)\\
			\hline
			\textbf{Step 1: }& Randomly split out a held-out validation dataset. The rest data are fed to the trained machine learning model to get the contribution matrix;\\
			\textbf{Step 2: }& Use Equation (\ref{eq1}) to split the initial node;\\
			\textbf{Step 3: }& Recursively partition the left and right child nodes $S_{L}$ and $S_{R}$ to get the full Tree $T_{0}$ until reaching to maximum tree depth or minimum number of data samples in one node;\\
			\textbf{Step 4: }& Use Equation (\ref{eq4}) to calculate the average split strength of each internal node of $T_{0}$. Iteratively remove internal nodes from the one with smallest split strength to get a series of nested tree, $\{T_{K}, T_{K-1}, ..., T_{k}, T_{k-1}, ...,T_{0}\}$;\\
			\textbf{Step 5: }& Use the held-out validation set and Equation (\ref{eq5}) and (\ref{eq6}) to calculate $G_{validation}(T_{k})$ for each of $\{T_{K}, T_{K-1}, ..., T_{k}, T_{k-1}, ...,T_{0}\}$. The one with largest  $G_{validation}(T_{k})$ is selected as the best sized interpretation tree;\\
			\hline            
		\end{tabular}
	\end{center}
	\caption{
		The complete algorithm to generate the interpretation tree.}
	\label{alg}
\end{table}

The full algorithm to generate the interpretation tree is described in Table \ref{alg}. Now we will move forward to demonstrate it on multiple datasets using various machine learning models.

\section{Experiment}
In this section, we will try to interpret different types of machine learning model on different tasks by representing them using the interpretation tree. First, We will apply the proposed Global Interpretation via Recursive Partitioning (GIRP) algorithm to a scene understanding deep learning model in computer vision. Second, we will try a text classification task and see what words are important to the random forest classifier. Finally, intensive care unit (ICU) mortality prediction using recurrent neural network from medical records demonstrates our approach on tabular data. Each of these cases is different in obtaining the contribution matrix. We will explain it in detail for each of them.

\subsection{Scene Understanding}
As many computer vision tasks greatly advanced by deep learning, scene understanding has breakthroughs in accuracy with the help of multi-million item labeled dataset and large scale deep neural networks \cite{zhou2016places}. However, as most successful neural network architectures for computer vision, scene understanding neural networks are not easily understandable because they are trained end-to-end and act in a black-box way. What's more, \cite{nguyen2015deep} shows that many popular network architectures are easily fooled. Though some workarounds are proposed to examine the evidence of neural predictions \cite{bau2017network}, they are at the single prediction level that could not be efficiently used when there are millions of training and testing data samples. Thus, people need a tool to extract the general rules contained in the model from the whole set of data. If these rules make sense to humans, we could trust the models more that they will generalize well in real world.

\begin{figure*}[!tb]
	\centering{} \includegraphics[width=160mm]{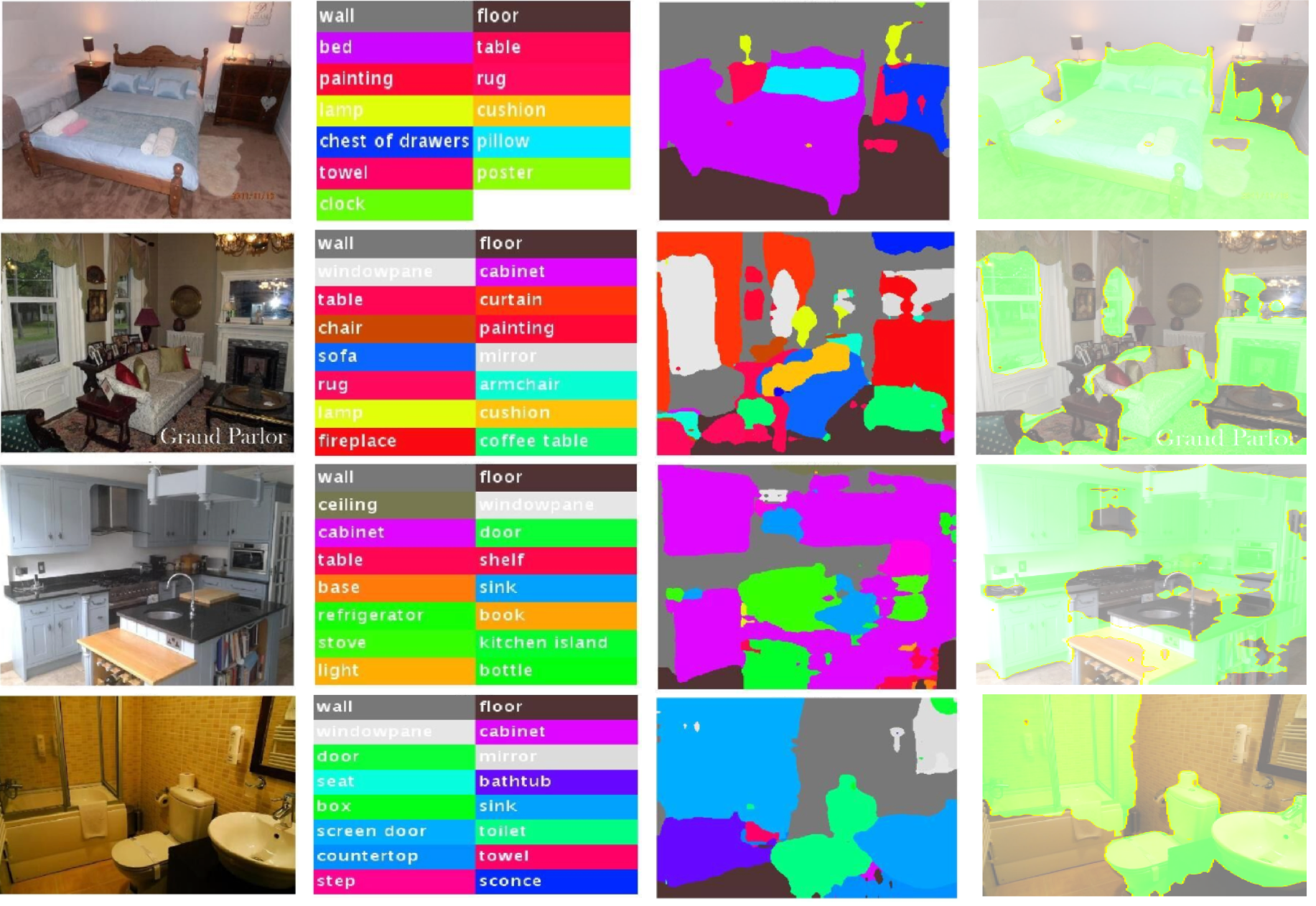} \caption{From row 1 to row 4, each row have one example image from the four categories of "bedroom", "living room", "kitchen", and "bathroom" in the MIT Place 365 scene understanding dataset \cite{zhou2016places}. The first column is the raw image. The second column shows that semantic categories that found by the semantic segmentation algorithm, dilated convolutional network \cite{yu2015multi} in the image. Column 3 shows actual semantic segmentation, which contains several superpixels for each image. Using the local prediction interpreter \cite{ribeiro2016should}, we could get the contribution of each superpixel, i.e., semantic category, to the predicted scores. Column 4 presents the important semantic superpixel (with highest contribution scores) that are highlighted green for corresponding ground truth category score respectively. For the "bedroom" image, the "bed" and "floor" superpixels are important. For the "living room" image, "sofa", "window pane", and "fireplace" are important. For the "kitchen" image, "cabinet" is the most important. Finally for the "bathroom" image, "toilet", "screen door" play the most important role. All these explanations seem to be reasonable to us human being.}
	\label{pic} 
\end{figure*}

In our demonstration, we will try to understand a deep residual network trained for scene understanding on the MIT Place 365 dataset \cite{he2016deep,zhou2016places}. To be specific, we will send images with ground truth label in "kitchen", "living room", "bedroom" and "bathroom" in the validation dataset, 100 images per category, to the trained model and collect the predicted probabilities for these four categories. To obtain the contribution matrix, we apply a scene parsing algorithm, dilated convolutional network \cite{yu2015multi,zhou2017scene}, to segment each image into semantically meaningful parts. Then we perturb each part with noise and re-evaluate the perturbed image in the scene understanding neural network for new prediction scores for the four categories. Using the varying scores, the contribution of each semantic part of the image can be calculated via a sparse linear regression model as the local model interpreter does \cite{ribeiro2016should}. Therefore, we could obtain the contribution of each semantic category to prediction scores of the four scene categories, which form the contribution matrix. Figure \ref{pic} describes this process more clearly.  

\begin{figure*}[t]
	\begin{minipage}[t]{\columnwidth}%
		\begin{center}
			\subfloat[Living room]{\includegraphics[width=90mm]{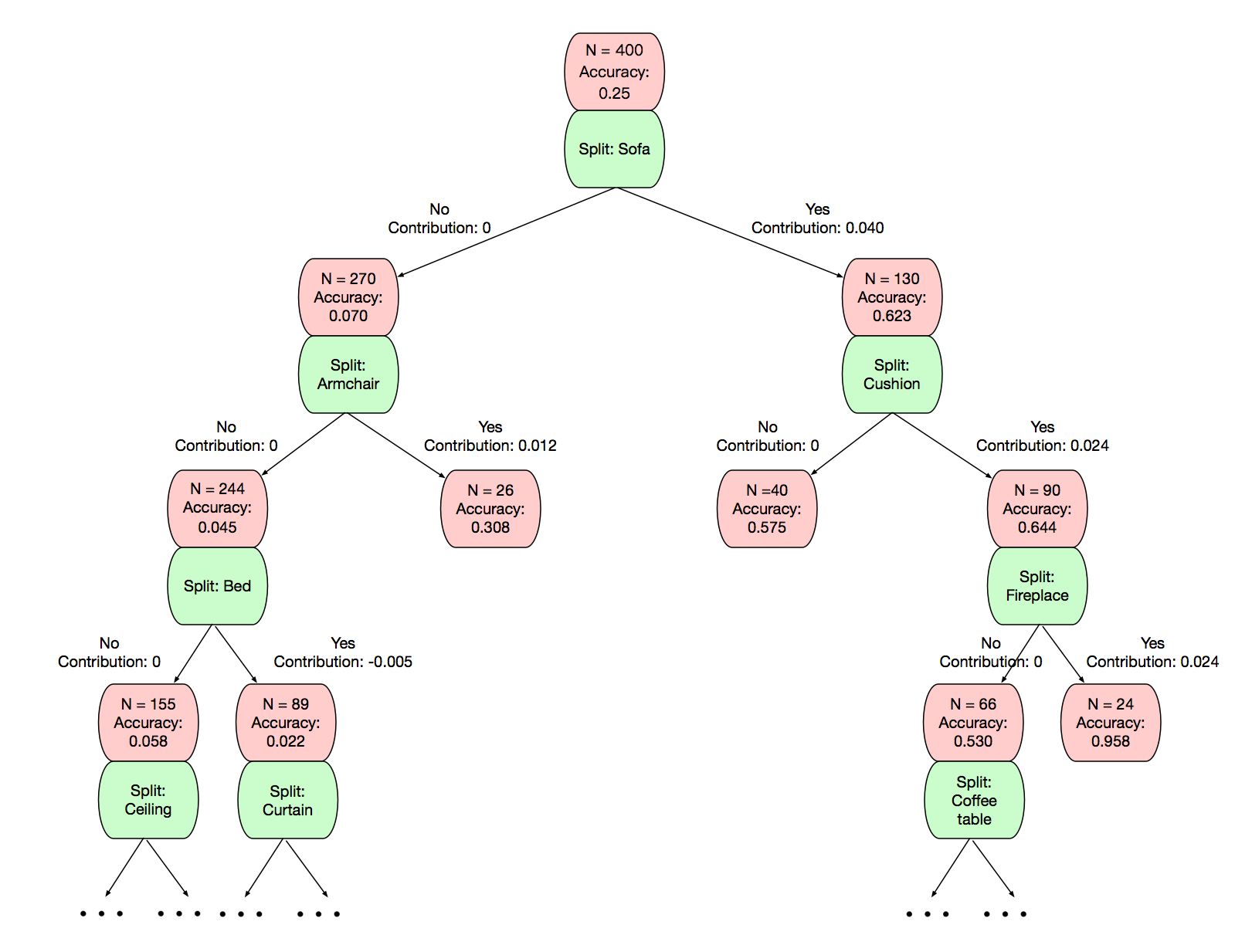} \label{qta} 
				
			}
			\par\end{center}%
	\end{minipage}\hfill{}%
	\begin{minipage}[t]{\columnwidth}%
		\begin{center}
			\subfloat[Kitchen]{\includegraphics[width=90mm]{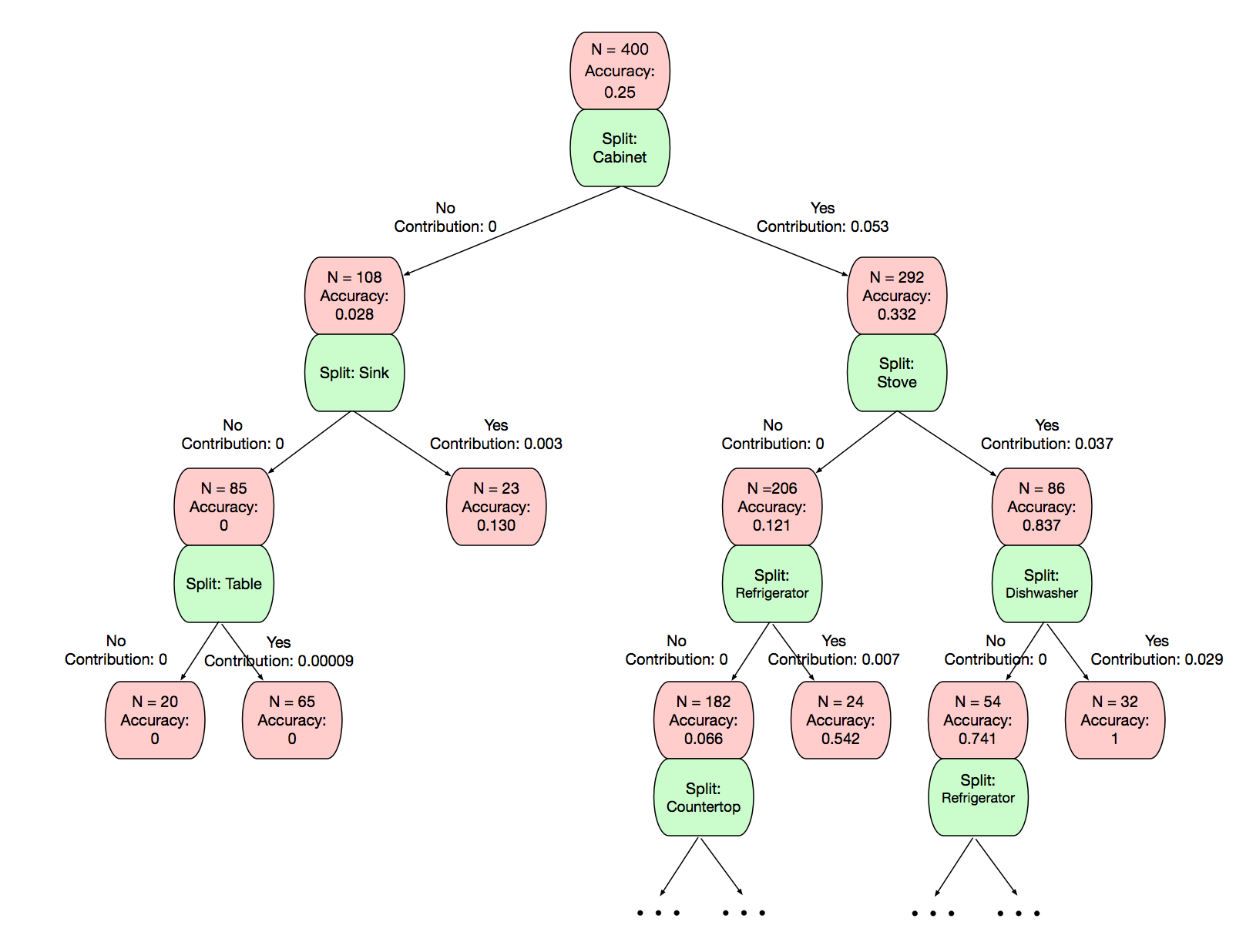}
				
			}
			\par\end{center}%
	\end{minipage}\hfill{}%
	
	\begin{minipage}[t]{\columnwidth}%
		\begin{center}
			\subfloat[Bedroom]{\includegraphics[width=90mm]{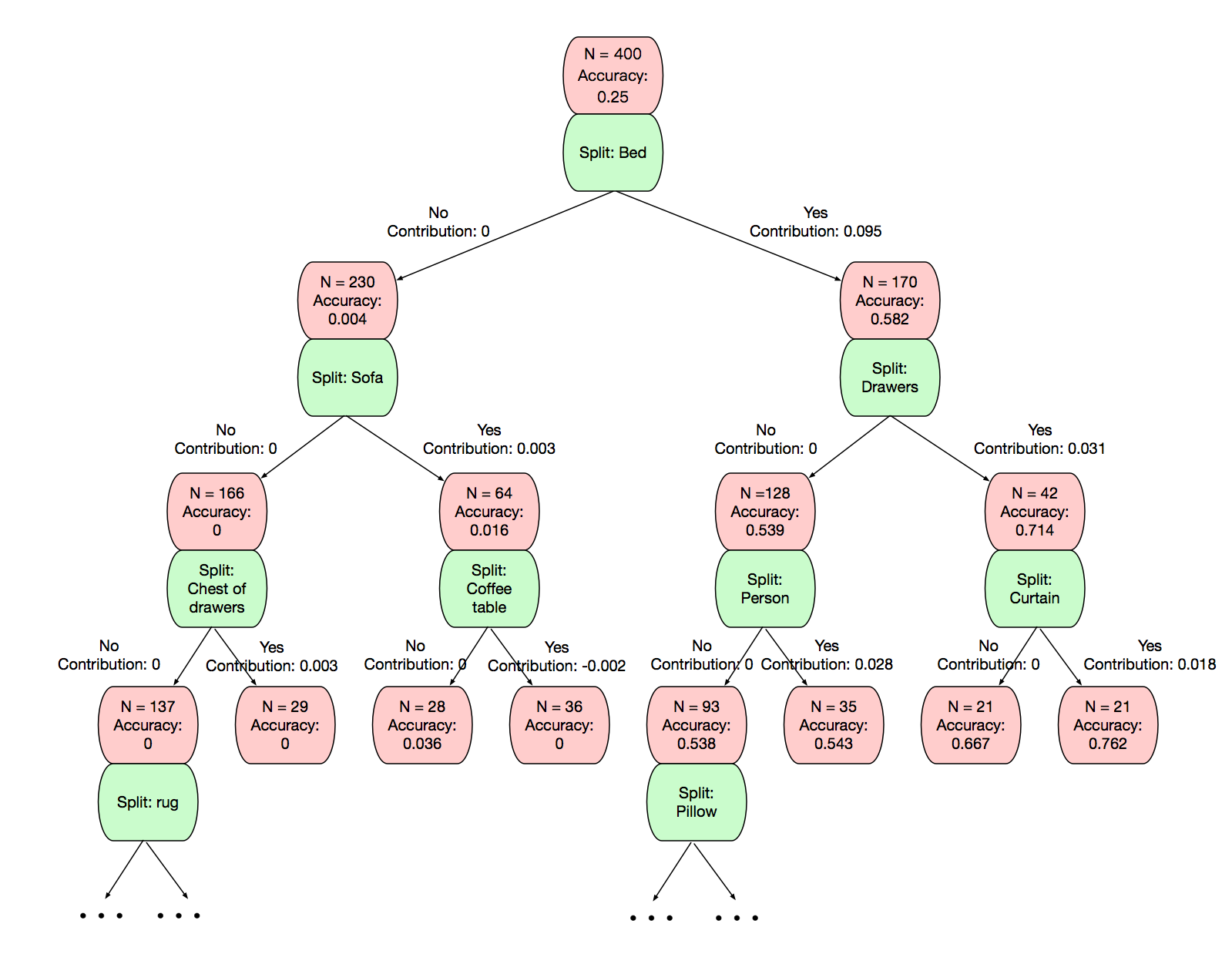} \label{qtd} 
				
			}
			\par\end{center}%
	\end{minipage}\hfill{}%
	\begin{minipage}[t]{\columnwidth}%
		\begin{center}
			\subfloat[Bathroom]{\includegraphics[width=90mm]{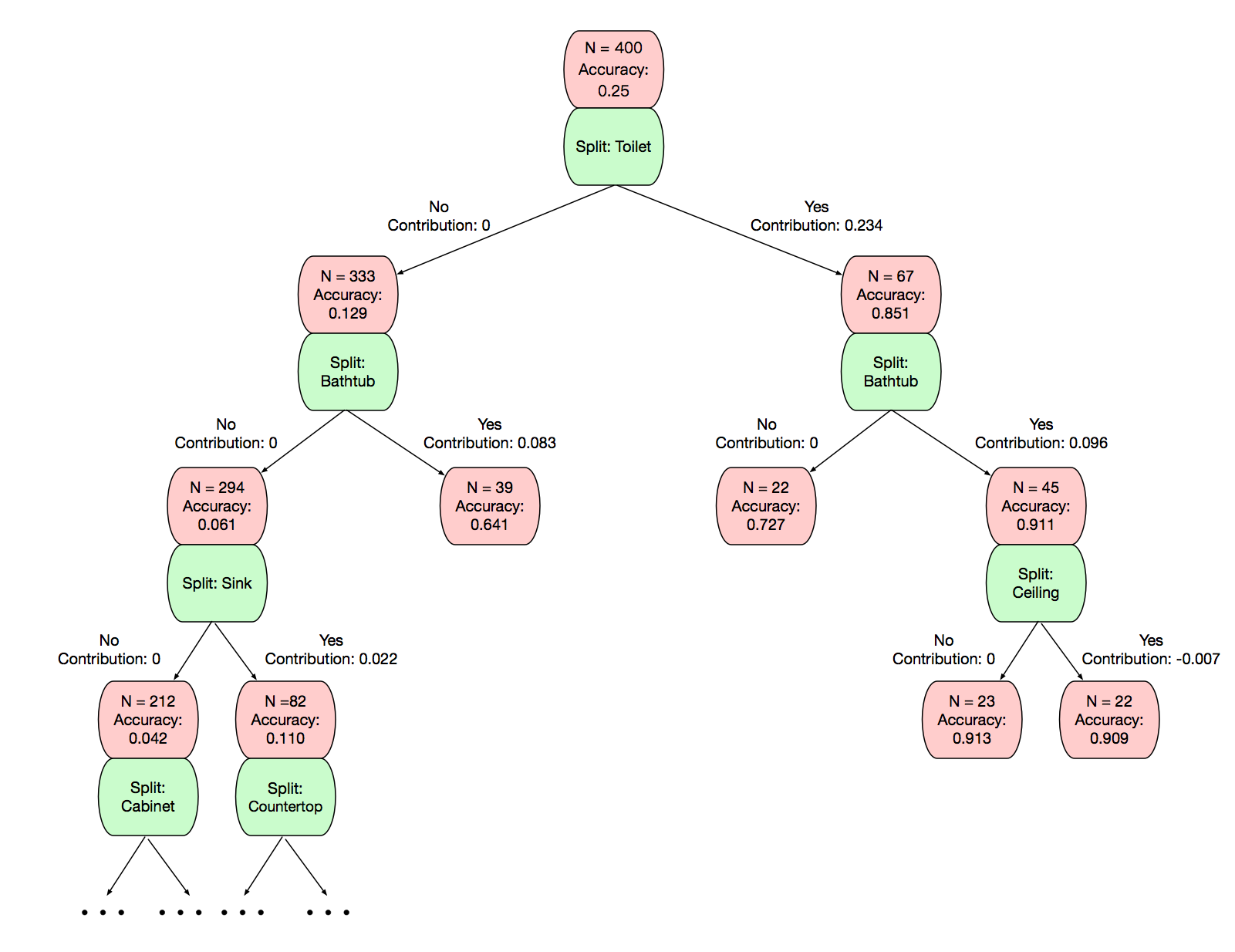} \label{qte} 
				
			}
			\par\end{center}%
	\end{minipage}\hfill{}%
	
	\caption{Interaction trees learned for scene categories "living room", "kitchen", "bedroom", and "bathroom".}
	\label{itcv}
\end{figure*}

After getting the contribution matrix that measures the importance of each semantic category for the scene categories for each image, we could run our Global Interpretation via Recursive Partitioning (GIRP) algorithm to generate the interpretation tree for each category, that is, "kitchen", "living room", "bedroom", and "bathroom". We set the maximum depth of the resulting tree to 100 and each node contains at least 20 images. The results are shown in Figure \ref{itcv}. Only the first four levels of the resulting trees are presented due to space limit. The actual best-sized tree is usually 5 to 10 levels in height. For each node in the interpretation tree, the numbers of images in the nodes are shown. The accuracy number measures what proportion of images are correctly identified as the ground truth category for each tree. The split variable for each node is also shown. The contribution number is the average contribution of the split variable in the left and right child node. From the trees, we can see that for "kitchen", "living room", "bedroom", and "bathroom" scenes, the model finds "cabinet", "sofa", "bed", and "toilet" are the most discriminative semantic categories, which does match our common sense. Besides, our approach also reveals some useful rules that the model is following. For example, the "sofa", "cushion", and "fireplace" combination achieve 0.958 in accuracy for identifying "living room", while the "cabinet", "stove", and "dishwasher" combination gets a perfect accuracy of 1 for "kitchen". All these findings increase our confidence in the black-box residual network based scene understanding deep learning model because it is picking the right important object in the scene to make decisions.

\subsection{Text Classification}
We now turn our attention to the text classification task. \cite{ribeiro2016should} reports that the text classifiers are picking up unreasonable words to discriminate articles related to "Christianity" from ones related to "Atheism" using a subset of the 20-newsgroups corpus. While they are showcasing this phenomenon by some randomly picked articles, we want to check if at the corpus level the model does use words unrelated to both concepts to classify articles. For this purpose, we use our proposed approach to generate an interpretation tree using words in the articles as features.

\begin{figure*}[t]
	\begin{minipage}[t]{\columnwidth}%
		\begin{center}
			\subfloat[Text Classification]{\includegraphics[width=90mm]{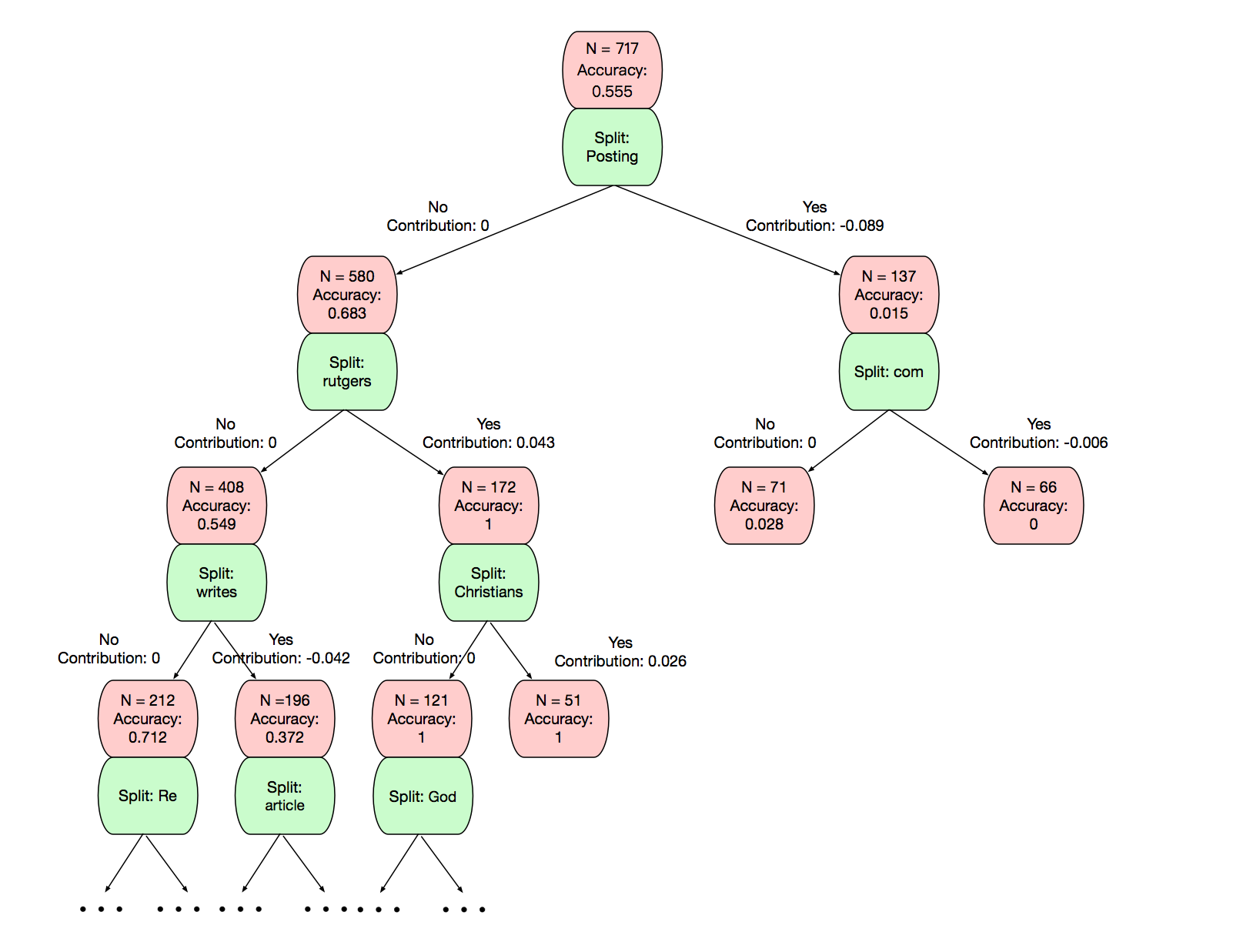} \label{txt} 
				
			}
			\par\end{center}%
	\end{minipage}\hfill{}%
	\begin{minipage}[t]{\columnwidth}%
		\begin{center}
			\subfloat[ICU Mortality]{\includegraphics[width=90mm]{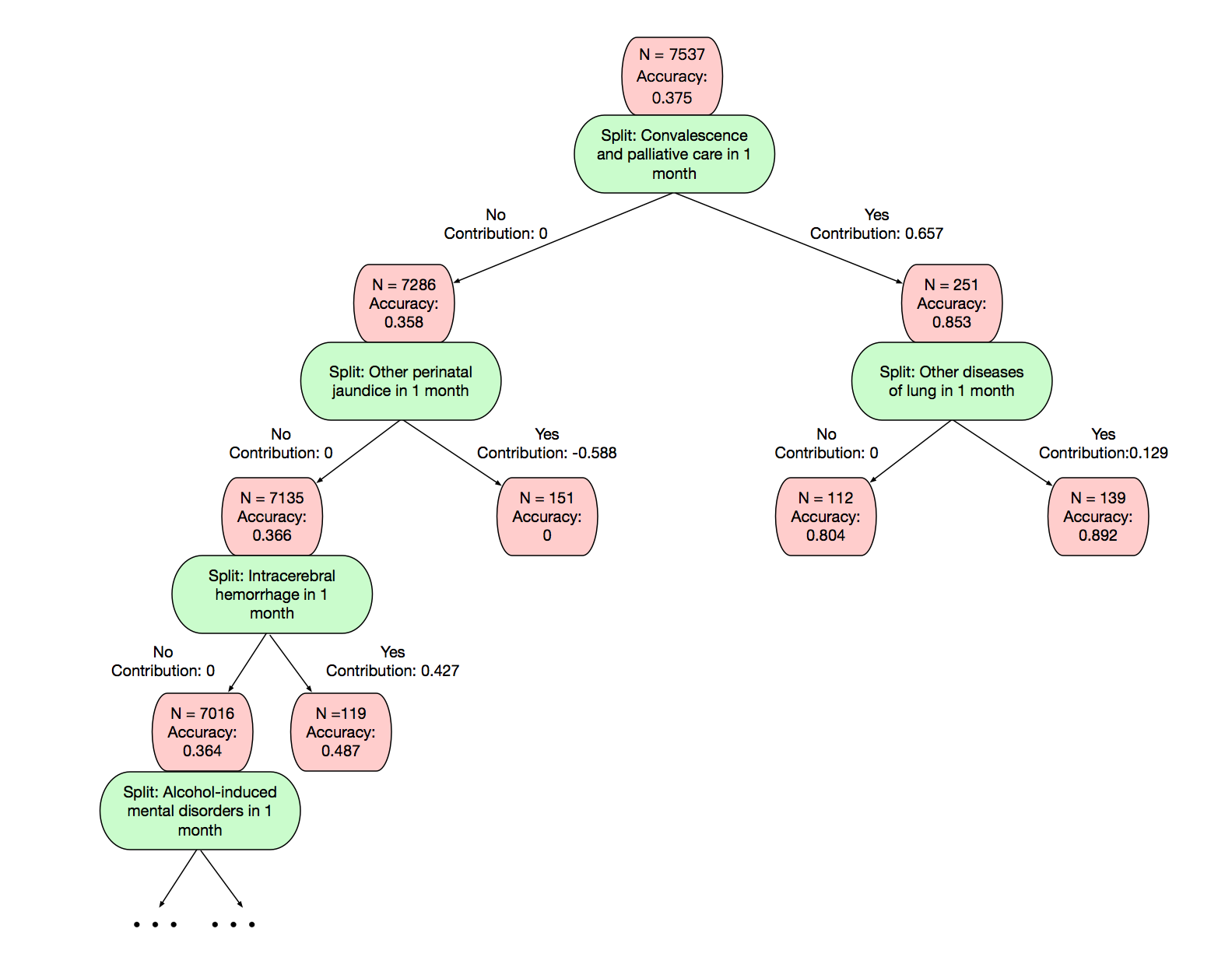}
				\label{icu}
			}
			\par\end{center}%
	\end{minipage}\hfill{}%
	
	\caption{\textbf{Left: Text Classification, }The interpretation tree to explain the random forest model classifying "Christianity" and "Atheism" related articles. Unfortunately, we can see from the tree that the model is picking up unreasonable words such as "Posting", "rutgers", and "com" as important features. We could expect bad generalizability of this model. \textbf{Right: ICU Mortality, }The interpretation tree to explain the recurrent neural network model predicting mortality in ICU using past medical records. The codes found by the algorithm are relevant to high or low risk of mortality.}
\end{figure*}

We train a random forest classifier \cite{liaw2002classification} with 500 trees that achieves 0.92 in accuracy on the test set to classify "Christianity" and "Atheism" articles. We use TF-IDF \cite{sparck1972statistical} vectorizer to transfer the article into vectors and then send them to the classifier. To get the contribution matrix for building the interpretation tree. we use the local interpreter \cite{ribeiro2016should} that removing each word from the articles one by one and monitoring the change in predicted score to do a regression for evaluating the contribution of each word. After running our Global Interpretation via Recursive Partitioning algorithm, we obtain an interpretation tree as shown in Figure \ref{txt}. Maximum tree depth is set to 100 and minimum number of data samples in each node is 50. The results show that most words found in the tree are not very related to concepts either of "Christianity" or “Atheism”, except "God" and "Christians" in the lower levels. The most important words pulled, "Posting", "Rutgers", and "com", look like just coincidental fake correlations captured by the model. \cite{ribeiro2016should} reports an imbalanced word frequency of these words in the two classes in the corpus. The model definitely overfits to these patterns and would not generalize well in classifying new articles. This finding implies that it would be a better practice to train robust text classification machine learning models from multiple corpora so that they are less likely overfitting to corpus specific features. In this text classification example, we show that GIRP and interpretation tree could be used to diagnose models that overfit to the data and reveal the incorrectly learned pattern.

\subsection{Tabular Data: Predicting ICU Mortality}\label{tb}
Structured tabular data widely exist in all kinds of relational databases to represent various types of events or transactions. Hospitals use standardized codes to record medical diagnosis, procedure and pharmacy codes. The MIMIC database \cite{johnson2016mimic} is this kind of medical database that contains intensive care unit (ICU) medical records and is publicly available. We apply the RETAIN algorithm \cite{choi2016retain} to the MIMIC database to predict mortality in the intensive care unit (ICU). RETAIN is a specifically designed interpretable recurrent neural network that can produce the contributions of past medical events to a predicted new event using the neural attention mechanism \cite{bahdanau2014neural}. However, due to the stochastic optimization process, \cite{yang2017machine} has shown that these contributions are not stable when the recurrent neural network model is re-trained by re-sampling training data. We apply our proposed Global Interpretation via Recursive Partitioning (GIRP) to see if the global interpretation of the RETAIN model is making sense when the local interpretations are unstable.

Past diagnosis codes are used to predict whether a patient will die in the ICU. The contribution of each diagnosis code is generated by RETAIN along with the prediction. For the convenience of interpretation, we aggregate the diagnosis codes to different time frames, though RETAIN predicts on continuous time series. We collect these contributions and organize them as the contribution matrix. Then it is sent to GIRP to generate the interpretation tree, which is shown in Figure \ref{icu}. Maximum tree depth is set to 100 and minimum number of data samples in each node is 100. We can see the most relevant diagnosis found by the algorithm is "convalescence and palliative care in 1 month" that indicates a mortality of 85.3\%. This is making sense because this diagnosis probably means most medical treatments are tried and the doctors can do nothing about the patients' situation. On the other hand, "other perinatal jaundice in 1 month" seems to be a big protective factor of death in the ICU. This is also reasonable because mostly jaundice is not life threatening but needs emergent care. For other codes we may not comment on the rationality because of the lack of health care knowledge. However, this figure may help doctors if it finds some relations between medical conditions and death in the ICU that are not well investigated before in the medical practice. In this way, our proposed method could potentially help discovering new important factors or interactions related to some outcome in complicated situations such as health care, which enables the black-box predictive models for knowledge discovery.

\section{Conclusion and Discussion}
In this paper, we propose a simple but effective method to interpret black-box machine learning models globally from local explanations of single data samples. Global interpretation is more refined than local explanations thus more efficient when used to diagnose the trained model or extract knowledge from it. We show that our Global Interpretation via Recursive Partitioning (GIRP) algorithm can represent many types of machine learning models in a compact manner. We demonstrate our algorithms using various kinds of machine learning models on different tasks. We have shown that the deep residual network is looking for the right object when classifying scenes. In contrast, in text classification, the interpretation tree indicates that the random forest classifier is focusing on wrong words to discriminate texts with different topics. Besides, the proposed method is also useful to extract decision rules from sophisticated models. Such rules are hidden in black-box models but are critical to know if we want to impact the outcome. We showcase this usage by extracting disease comorbidities leading to high mortality in intensive care unit. In conclusion, our method helps people understand machine learning models efficiently, making it easier to check if the model is behaving reasonably and make use of the knowledge it discovers.

However, the proposed method is limited in several ways. First, we lack a quantitative measure of the fidelity of the interpretation tree to the original explained machine learning model. Though the interpretation tree is directly developed from contributions generated from the original model, we lose some details when we extract the general rules. We don't know how important these details are to the high predictability. Second, though we present the split strength as a measure of variable importance in the interpretation tree, the confidence for this measure is unknown. Linear methods are popular in evidence based studies partially because it is easier for confidence interval estimation. Due to the complexity of underlying probabilistic distributions for sophisticated machine learning methods, it is difficult to estimate confidence intervals for the split strengths in the interpretation tree. Finally, the proposed method needs a contribution matrix as an input which is difficult to obtain when feature representation from input variables is not well established such as speech recognition and computer vision. For example, in many vision tasks, even local visual explanations are available using image segmentation \cite{ribeiro2016should,yang2016supervoxel,yang2018visual}, it is difficult to aggregate them to high level visual features. This is closely connected to the broader problem and theories such as the information bottleneck \cite{tishby2000information,tishby2015deep} of understanding how machine learning models identify important high level features and ignore the noises. Additionally, even the explicit feature representation is available to form the contribution matrix, group effects of features are not well captured in the current method. Mechanisms similar to group LASSO \cite{yuan2006model} could be added to solve this problem.

Each of the limitations mentioned points to a good direction for future work. We want to quantify the fidelity of the interpretation tree to the explained original model. We are considering setting up bootstrapping methods \cite{efron1979bootstrap} for confidence interval estimation as direct probabilistic distribution estimation is difficult. At last, some representation learning methods could be incorporated into the algorithm when the contribution matrix is difficult to obtain. All these pose exciting challenges for making machine learning more transparent for human beings.
\bibliographystyle{IEEEtran}
\bibliography{IEEEabrv,icdm17}
	
\end{document}